\documentclass{article}
\usepackage{ijcai25}

\usepackage{times}
\usepackage{soul}
\usepackage{url}
\usepackage[hidelinks]{hyperref}
\usepackage[utf8]{inputenc}
\usepackage[small]{caption}
\usepackage{graphicx}
\usepackage{amsmath}
\usepackage{amsthm}
\usepackage{booktabs}
\usepackage{algorithm}
\usepackage{algorithmic}
\usepackage[switch]{lineno}
\usepackage{bm}
\usepackage[UKenglish]{babel}
\usepackage[babel]{microtype}
\usepackage{amssymb}
\usepackage{mathtools}
\usepackage{nicefrac}
\usepackage{listings}
\usepackage{lstautogobble}
\usepackage{booktabs}
\usepackage{array}
\newcolumntype{R}[1]{>{\raggedleft\arraybackslash}p{#1}}
\newcolumntype{L}[1]{>{\raggedright\arraybackslash}p{#1}}
\newcolumntype{C}[1]{>{\centering\arraybackslash}p{#1}}
\usepackage[appendix=append,bibliography=common]{apxproof}
\newcommand{\mycommentstyle}{\normalfont\itshape}
\newcommand{\lat}{L}
\newcommand{\set}{\mathcal{L}}
\DeclareMathOperator{\glb}{\wedge}
\DeclareMathOperator{\lub}{\vee}
\DeclareMathOperator*{\bigGlb}{\bigwedge}
\DeclareMathOperator*{\bigLub}{\bigvee}
\DeclareMathOperator{\op}{O}
\DeclareMathOperator{\lfp}{lfp}
\newcommand{\biLat}{\lat^2}
\newcommand{\biSet}{\set^2}
\newcommand{\cons}{c}
\newcommand{\consSet}{\set^\cons}
\DeclareMathOperator{\aOp}{\mathcal{O}}
\DeclareMathOperator{\aOpL}{\aOp_l}
\DeclareMathOperator{\aOpU}{\aOp_u}
\DeclareMathOperator{\stable}{\mathcal{S}}
\newcommand{\ult}{u}
\DeclareMathOperator{\uOp}{\aOp^\ult}
\newcommand{\mon}{M}
\newcommand{\0}{\bm{0}}
\newcommand{\1}{\bm{1}}
\newcommand{\sem}{S}

\newcommand{\prog}{P}
\DeclareMathOperator{\lpif}{\mathtt{:-}}
\DeclareMathOperator{\interp}{I}
\newcommand{\atoms}{\mathcal{A}_\prog}
\newcommand{\interps}{IS_\prog}
\newcommand{\inters}{\bigcap}
\newcommand{\semModels}{MS_\prog}
\newcommand{\minSemModel}{M_\prog^s}

\DeclareMathOperator{\imCon}{T_\prog}
\DeclareMathOperator{\leqNat}{\leq_N}
\DeclareMathOperator{\leqC}{\leq_C}
\DeclareMathOperator{\leqPrec}{\leq_p}
\DeclareMathOperator{\leqVal}{\leq_v}
\DeclareMathOperator{\botSem}{\bot_\leq}
\DeclareMathOperator{\botInterp}{\bot_\preceq}
\DeclareMathOperator{\botPrec}{\bot_{\leqPrec}}

\DeclareMathOperator{\botNat}{\bot_{\leqNat}}
\DeclareMathOperator{\topSem}{\top_\leq}

\DeclareMathOperator{\topNat}{\top_{\leqNat}}
\DeclareMathOperator{\negSem}{not}

\DeclareMathOperator{\uImCon}{\mathcal{T}_\prog^\ult}
\DeclareMathOperator{\fImCon}{\mathcal{T}_\prog^f}
\DeclareMathOperator{\fImConL}{\fImCon_l}
\newcommand{\consInterps}{\interps^\cons}
\newcommand{\biInterps}{\interps^2}

\theoremstyle{plain}
\newtheoremrep{theorem}{Theorem}
\newtheoremrep{lemma}{Lemma}
\newtheoremrep{corollary}{Corollary}

\theoremstyle{definition}
\newtheorem{definition}{Definition}

\theoremstyle{remark}
\newtheorem*{note}{Note}

\newtheorem{example}{Example}

\title{A Unifying Framework for Semiring-Based Constraint Logic Programming With Negation (full version)}

\author{
Jeroen Spaans$^1$
\And
Jesse Heyninck$^{1,2}$\\
\affiliations
$^1$Open Universiteit, The Netherlands\\
$^2$University of Cape Town, South Africa\\
\emails
\{jeroen.spaans, jesse.heyninck\}@ou.nl
}

\begin{document}

\maketitle

\begin{abstract}
    Constraint Logic Programming (CLP) is a logic programming formalism used to solve problems requiring the consideration of constraints, like resource allocation and automated planning and scheduling.
    It has previously been extended in various directions, for example to support fuzzy constraint satisfaction, uncertainty, or negation, with different notions of semiring being used as a unifying abstraction for these generalizations.
    None of these extensions have studied clauses with negation allowed in the body. 
    We investigate an extension of CLP which unifies many of these extensions and allows negation in the body.
    We provide semantics for such programs, using the framework of approximation fixpoint theory, and give a detailed overview of the impacts of properties of the semirings on the resulting semantics.
    As such, we provide a unifying framework that captures existing approaches and allows extending them with a more expressive language.
\end{abstract}

\begin{toappendix}
\section{Additional Results}
\subsection{Additional results on semirings}
\label{Appendix:AdditionalResults:Semirings}
In this section, we make some additional observations on semirings.
\begin{lemmarep}
    \label{lem:incompleteness, idempotent semiring and c-semiring}
    Not every idempotent semiring, nor every c-semiring, is complete.
\end{lemmarep}
\begin{nestedproof}
    Consider the semiring $\langle \mathbb{Q}^+\cup\{\infty\}, \min, +, \infty , 0\rangle$.
    It can be verified that this is a c-semiring.
    However, it is not complete, as $\min(\{x\in \mathbb{Q}\mid x\geq \sqrt{2}\})=\sqrt{2} \notin \mathbb{Q}^+\cup\{\infty\}$.
\end{nestedproof}

\subsection{Traditional models}
\label{Appendix:AdditionalResults:TraditionalModels}
In this appendix we make some observations on what we call \emph{traditional models}, and compare these to the semiring models introduced in the paper.
\begin{note}
    In this section, as in Section \ref{sec:semiring-based constraint logic programming} and unless otherwise specified, we assume $\sem = \langle \set, +, \times, \0, \1 \rangle$ to be a commutative semiring such that $\langle \set, \leq \rangle$ is a partial order, and that $\prog$ is a PSCLP($\sem$).
\end{note}
Traditionally, models rely on a notion of satisfaction of clauses.
\begin{definition}[Clause Satisfaction]
    \label{def:clause satisfaction}
    $\interp \in \interps$ \emph{satisfies} a clause $H \lpif B$ in $\prog$ under $\leq$ if $\interp(B) \leq \interp(H)$.\footnote{When it is clear from context, the ordering under which a clause is satisfied is left implicit.}
\end{definition}
\begin{definition}[Traditional Model]
    $\interp \in \interps$ is a \emph{traditional model} of $\prog$ if it satisfies all clauses of $\prog$.
\end{definition}
The notions of model intersection and minimal model for traditional models are defined analogously to those for semiring models.

As with semiring models, we have that the intersection of a set of traditional models is also a traditional model.
\begin{lemmarep}
   ~\cite{bistarelliSemiringbasedContstraintLogic2001}
    \label{lem:traditional model intersection}
    If $\sem$ is ordered by the complete lattice $\langle \set, \leq \rangle$, and $M$ is a set of traditional models for $\prog$, then $\inters M$ is a traditional model for $\prog$.
\end{lemmarep}
\begin{nestedproof}
    Let $\prog$ be a PSCLP based on a semiring $\sem = \langle \set, +, \times, \0, \1 \rangle$, that is ordered by a complete lattice $\langle \set, \leq \rangle$.
    Let $M$ be a set of traditional models for $\prog$.
    Since each $\interp \in M$ is a traditional model, we have for each $\interp \in M$ and every clause $H \lpif B$ in $\prog$ that $\interp (B) \leq \interp (H)$ holds.
    We show that, for all $H \lpif B$ in $\prog$, $\inters M (B) \leq \inters M (H)$ holds too.

    We consider first the interpretation of a body and show that ($\dagger$): $\inters M (B) \leq \interp (B)$ holds for any $H \lpif B \in \prog$ and each $\interp \in M$.
    Take any $H \lpif B_1,\ldots,B_n \in \prog$.
    We have $\interp (B_1,\ldots,B_n) = \prod_{i=1}^n B_i$ and $\inters M (B_1,\ldots,B_n) = \prod_{i=1}^n \{ \bigGlb\{\interp(B_i) \mid \interp \in M \} \}$.
    By definition of glb, we have that $\bigGlb\{\interp(B_i) \mid \interp \in M \} \leq \interp (B_i)$ for any of the $B_i$ occurring in the body of our clause.
    From here, since $\times$ is monotone, we get that $\prod_{i=1}^n \{ \bigGlb\{\interp(B_i) \mid \interp \in M \} \} \leq \prod_{i=1}^n B_i$, and, thus, that $\inters M (B) \leq \interp (B)$ holds for any $H\lpif B \in \prog$.
    This concludes the proof for $\dagger$.

    For the interpretation of a head---heads being singular atoms---we have that ($\ddagger$): $\inters M (H) = \bigGlb \{\interp (H) \mid \interp \in M \}$ holds for any $H \lpif B \in \prog$.

    Now take any traditional model $\interp \in m$ and any clause $H \lpif B$ in $\prog$ and recall that, since $\interp$ is a traditional model, $\interp (B) \leq \interp (H)$ holds.
    By $\dagger$ we have that $\inters M (B) \leq \interp (B)$.
    By transitivity of $\leq$, we now get that $\inters M (B) \leq \interp(H)$ --- i.e. $\inters M (B)$ is a lower bound of $\{\interp(H) | \interp \in M \}$.
    Finally, $\ddagger$ and the definition of \emph{greatest} lower bound give us that $\inters M (B) \leq \inters M (H)$, concluding our proof.
\end{nestedproof}

In Section \ref{sec:semiring-based constraint logic programming}, we found that the prefixpoints of $\imCon$ coincide with semiring models or $\prog$.
We do not generally have such a coincidence between traditional models of $\prog$ and the prefixpoints of $\imCon$.
\begin{example}[Traditional Models and Prefixpoints of $\imCon$]
    \label{ex:traditional models and prefixpoints of tp}
    To illustrate that traditional models and prefixpoints do not necessarily coincide, we present a simple counterexample wherein a traditional model is not a prefixpoint of $\imCon$
    Consider a PSCLP($\mathbb{N}_\infty$) consisting of the clauses $H \lpif B_1$ and $H \lpif B_2$.
    Take some interpretation $\interp \in \interps$ s.t. $\interp(H) = \interp(B_1) = \interp(B_2) = 5$.
    $\interp$ satisfies both clauses in $\prog$, so $\interp$ is a traditional model of $\prog$.
    Applying $\imCon$, we have $\imCon(\interp)(H) = \interp(B_1) + \interp(B_2) = 5 + 5 = 10$---$\imCon(\interp)(H) > \interp(H)$, so $\imCon(\interp) \npreceq \interp$.
    $\interp$ is a traditional model of $\prog$, but not a prefixpoint of $\imCon$.
\end{example}
In our example, there are two clauses defining the predicate in $H$.
$\imCon$, considering all clauses defining a predicate, adds the interpretations of these clauses together.
A traditional model, in contrast, does not consider all clauses defining a predicate together, instead looking at each clause individually.

Because we cannot establish the equivalence between traditional models and prefixpoints of $\imCon$, we also cannot establish an equivalence between the minimal traditional model and the least fixpoint of $\imCon$.
This means that traditional models would not generally lead to a semantics that is equivalent to our fixpoint semantics.

We note that traditional models and semiring models coincide for c-semirings.
\begin{lemmarep}
    Let $\sem$ be a c-semiring. If $\interp$ is a traditional model, it is also a semiring model.
\end{lemmarep}
\begin{nestedproof}
We first show that ($\dagger$): for every rule $ H \lpif B_1, \ldots, B_n \in \prog$, there is some $ H \lpif C_1, \ldots, C_m \in \prog$ s.t.: 
 $\sum\{\prod_{i=1}^{n} \interp(B_i)\mid H \lpif B_1, \ldots, B_n \in \prog\} = \prod_{i=1}^{m} \interp(C_m)$. Indeed, we first observe that, as $\sum\{\prod_{i=1}^{n} \interp(B_i) |\mid H \lpif B_1, \ldots, B_n \in \prog\}=
\sum\{\prod_{i=1}^{n} \interp(B_i) \mid H \lpif B_1, \ldots, B_n  \in \prog \setminus \{ H \lpif C_1, \ldots, C_m   \}\} +\prod_{i=1}^{m} \interp(C_m)$, as a C-semiring has an idempotent $+$, $\sum\{\prod_{i=1}^{n} \interp(B_i) \mid H \lpif B_1, \ldots, B_n \in P\}\geq  \prod_{i=1}^{m} \interp(C_m)$. 
Furthermore, as $\sum\{\prod_{i=1}^{n} \interp(B_i) \mid H \lpif B_1, \ldots, B_n \in \prog\}= \bigvee \{ \prod_{i=1}^{n} \interp(B_i)    \mid  H \lpif B_1, \ldots, B_n \in \prog\}$ (as shown by \cite{bistarelliSemiringbasedContstraintLogic2001}), this concludes the proof of $\dagger$.

Suppose now $\interp$ is a model of $\prog$, i.e.\ for every $ H \lpif B_1, \ldots, B_n \in \prog$, $ \prod_{i=1}^{n} \interp(B_i)\leq \interp(H)$. With $\dagger$, there is some $ H \lpif C_1, \ldots, C_m \in \prog$ s.t.\ $\sum\{\prod_{i=1}^{n} \interp(B_i)\mid H \lpif B_1, \ldots, B_n \in \prog\} = \prod_{i=1}^{m} \interp(C_m)$. Thus, $\sum\{\prod_{i=1}^{n} \interp(B_i)\mid H \lpif B_1, \ldots, B_n \in \prog\} \leq \interp(H)$, which means that $M$ is a semiring model.
\end{nestedproof}
\end{toappendix}
\section{Introduction}

Constraint logic programming (CLP) facilitates solving problems requiring the consideration of constraints, and is well-established as a fruitful formalism over the last decades.
To further increase its expressivity, its language has been extended to take into account information about a problem domain of a more quantitative nature, e.g.\ by allowing fuzzy constraint satisfaction or taking into account scores or penalties.
Due to the wide variety of such possible extensions, several works have sought to unify these approaches using the algebraic concept of a \emph{semiring} \cite{bistarelliSemiringbasedContstraintLogic2001,khamisConvergenceDatalogPre2023}.
Even though these works represent a big step forward in terms of unification, they introduce their own assumptions on semirings, which makes it hard to compare these unifying approaches.
Furthermore, they restrict attention to positive logic programs, i.e.\ programs consisting of rules without negation in the body.
This is unfortunate, as it is well-known from e.g.\ answer set programming that negation increases both the conceptual and computational expressivity.
However, it also introduces non-monotonicity of the corresponding immediate consequence operator, leading to intricacies in the definition of semantics.

\emph{Approximation fixpoint theory} (AFT)~\cite{deneckerApproximationsStableOperators2001} is a unifying framework for the definition and study of semantics of non-monotonic formalisms.
It is a purely algebraic theory which was shown to unify the semantics of, among others, logic programming, default logic, and autoepistemic logic.
Due to the algebraic constructions defined in AFT, it often suffices to define a so-called approximating operator to derive a whole class of well-behaved semantics, including the so-called stable and well-founded semantics.
 
In this paper, we apply AFT to unify and generalize existing approaches to semiring-based constraint logic programming. In more detail, the contributions of this paper are the following:
(1) we unify the approaches of \cite{bistarelliSemiringbasedContstraintLogic2001,khamisConvergenceDatalogPre2023} in the framework of AFT, making clear the different assumptions used in these approaches and how they affect the behaviour of the corresponding semantics, 
(2) we generalize the language of semiring-based constraint logic programming to allow for negation in the body of rules, and
(3) use AFT to define and study semantics for such programs, introducing among others the stable and well-founded semantics, and show how they generalize both existing approaches to (positive) SCLPs and normal logic programs (nlps).
 
\noindent{\bf Outline of the Paper}: In Section \ref{sec:general background} we provide background on lattice theory, AFT and semirings. Section \ref{sec:families of semirings and their orderings} considers properties of semirings and derived orders. Section \ref{sec:semiring-based constraint logic programming} reconstructs existing semantics for SCLPs with lifted assumptions, and in Section \ref{sec:generalized negation for semiring-based constraint logic programming}, we introduce the syntax and semantics of negation, whereas in Section \ref{sec:normal semiring-based constraint logic programming in approximation fixpoint theory} the semantics for programs with negation are studied. We conclude with regard to related work in Section \ref{sec:related:work}. \section{General Background}
\label{sec:general background}
In this section, we introduce the necessary preliminaries on lattice theory (Section \ref{sec:lattice:theory}), approximation fixpoint theory (Section \ref{ssec:general background approximation fixpoint theory}) and semirings (Section \ref{sec:semirings}).

\subsection{Lattice Theory}\label{sec:lattice:theory}

A \emph{partially ordered set (poset)} $\lat$ is an ordered pair $\langle \set,\leq \rangle$, where $\leq\subseteq \set\times\set$ is a reflexive, antisymmetric and transitive relation. 
Given a poset  $\lat = \langle \set,\leq \rangle$ with $S \subseteq \set$, 
A \emph{lower bound} (resp. \emph{upper bound}) of $S$ is an element $x \in \set$ such that $x \leq s$ (resp. $s \leq x$) for all $s \in S$.
A lower bound $y$ of $S$ is the unique \emph{greatest lower bound} (glb) of $S$, denoted $\bigGlb S$, if for every lower bound $x$ of $S$ in $\set$ we have that $x \leq y$, and similarly for the  \emph{least upper bound} (lub). We also denote   $x \glb y$ by $\bigGlb \{x,y\}$ and $x \lub y$ by  $\bigLub \{x,y\}$.

A \emph{lattice} is a partially ordered set such that each pair of elements has a greatest lower bound and a least upper bound. A lattice is \emph{bounded} if there are $\bot, \top \in \set$ such that for any $x \in \set$, $\bot \leq x \leq \top$ holds, and it is complete if every subset of $\set$ has a greatest lower bound and a least upper bound. $\botSem$ denotes a complete lattice's least element, and $\topSem$ its greatest element.
While every complete lattice is a bounded lattice, it should be noted that the converse is not true~\cite{blythLatticesOrderedAlgebraic2005}.

We will study operators $\op: \set \to \set$, which are $\leq$\emph{-monotone} if for all $x \leq y$ we have that $\op(x) \leq \op(y)$, and $\leq$\emph{-antimonotone} if for all $x \leq y$ we have that $\op(y) \leq \op(x)$.\footnote{When the ordering clear, we simply say $\op$ is \emph{(anti)monotone}.}
This work also concerns commutative binary operators $\op: \set \times \set \to \set$.
Such operators are called (anti)monotone if they are monotone (resp. antimonotone) in either operand.
That is, we fix one operand at a time and consider the (anti)monotonicity of $x \mapsto \op(x,y)$ and $y \mapsto \op(x,y)$ separately.
An element $x \in \set$ is a \emph{prefixpoint} of $\op$ if $\op(x) \leq x$, a \emph{fixpoint} of $\op$ if $\op(x) = x$, and a \emph{postfixpoint} of $\op$ if $x \leq \op(x)$.   Should it exist, $\lfp_\leq(\op)$ denotes the $\leq$\emph{-least fixpoint} of $\op$.\footnote{If the ordering is clear, we simply write $\lfp(\op)$.}

Thanks to Theorem 5.1 of Cousot and Cousot~\cite{cousotConstructiveVersionsTarski1979}, we have a constructive characterization of this least fixpoint \cite{tarskiLatticetheoreticalFixpointTheorem1955}.
For this characterization, we first define the ordinal powers of a function  $\op: \set \to \set$ by defining $\op^0 (x) = x$,  $\op^{\alpha+1}(x) = \op(\op^\alpha(x))$ for a successor ordinal $\alpha$, and $\op^\alpha(x) = \bigLub_{\beta < \alpha}\op^\beta(x)$ for a limit ordinal $\alpha$.
The least fixpoint of a monotone operator on a complete lattice can be constructed by applying its ordinal powers starting from $\botSem$:
\begin{theorem}[\cite{cousotConstructiveVersionsTarski1979}]
    \label{thm:cousot and cousot construction}
    Given a complete lattice $\lat = \langle \set,\leq \rangle$ and a $\leq$-monotone operator $\op: \set \to \set$, there is an ordinal $\alpha$ such that $\op^\alpha(\botSem)$ is the least fixpoint of $\op$, which coincides with $\op$'s least prefixpoint.
\end{theorem}

\subsection{Approximation Fixpoint Theory}
\label{ssec:general background approximation fixpoint theory}
In this section, we recall \emph{approximation fixpoint theory} (AFT) \cite{deneckerApproximationsStableOperators2001}, which offers techniques for approximating the fixpoints of a possibly non-monotonic operator. 

Given a lattice $\lat = \langle \set, \leq \rangle$, a \emph{bilattice} is the structure $\biLat = \langle \biSet, \leqPrec, \leqVal \rangle$, where $\biSet = \set \times \set$, and for every $x_1,y_1,x_2,y_2 \in \set$, $(x_1,y_1) \leqPrec (x_2,y_2)$ if $x_1 \leq x_2$ and $y_2 \leq y_1$, and  $(x_1,y_1) \leqVal (x_2,y_2)$ if $x_1 \leq x_2$ and $y_1 \leq y_2$. Intuitively, bilattice elements $(x,y) \in \biSet$ approximate elements in the interval $[x,y] = \{z \in \set : x \leq z \leq y \}$. If $x\leq y$, we call $(x,y)$  \emph{consistent}, and $\consSet$ is the set  of all consistent pairs.
We call elements $(x,x) \in \consSet$ \emph{exact}, and note that the set of all exact elements constitutes an embedding of $\set$ in $\biSet$. We also use projection functions $(x,y)_1 = x$ and $(x,y)_2 = y$.

An operator $\aOp:\biSet \to \biSet$ is an \emph{approximator of} $\op: \set \to \set$ if it is $\leqPrec$-monotone and, for any $x \in \set$, $\aOp (x,x) = (\op(x),\op(x))$.
Approximating operators $\aOp$ can be thought of as combinations of two separate operators $(\aOp(\cdot,\cdot))_1$ and $(\aOp(\cdot,\cdot))_2$ calculating, respectively, a lower and upper bound for the value of  the approximated operator $\op$. We denote $(\aOp(\cdot,\cdot))_1$ and $(\aOp(\cdot,\cdot))_2$ as $\aOpL(x,y)$ and $\aOpU(x,y)$ respectively.
Specifically, we have $\aOpL (\cdot,y) = \lambda x.\aOpL(x,y)$ (i.e., $\aOpL (\cdot,y)(x) = \aOpL(x,y)$) and $\aOpU (x,\cdot) = \lambda y.\aOpU(x,y)$.

Given a complete lattice $\langle \set,\leq \rangle$ and an approximating operator $\aOp : \biSet \to \biSet$, the \emph{stable operator} for $\aOp$ is $\stable(\aOp)(x,y)=(\lfp(\aOpL(\cdot,y)),\lfp(\aOpU(x,\cdot)))$.
For any $\leqPrec$-monotone operator $\aOp$ on $\biSet$, the fixpoints of $\stable(\aOp)$ are $\leqVal$-minimal fixpoints of $\aOp$~\cite{deneckerApproximationsStableOperators2001}.
Also, $\aOpL(\cdot,y)$ and $\aOpU(x,\cdot)$ are $\leq$-monotone operators ---guaranteeing $\stable(\aOp)$ is well-defined, and $\stable(\aOp)$ is $\leqPrec$-monotone~\cite{deneckerApproximationsStableOperators2001}.

Given an approximating operator $\aOp : \biSet \to \biSet$.  $(x,y)$ is:  the \emph{Kripke-Kleene (KK) fixpoint} of $\aOp$ if $(x,y) = \lfp_{\leqPrec}(\aOp)$;
a \emph{stable} fixpoint of $\aOp$ if $(x,y)= \stable(\aOp)(x,y) $;
the \emph{well-founded (WF) fixpoint} of $\aOp$ if $(x,y)=\lfp_{\leqPrec}(\stable(\aOp))$.

A lattice operator may permit multiple approximating operators, but
Denecker et al.~\citeyear{deneckerUltimateApproximationIts2004} define the most precise approximating operator $\uOp$ approximating an operator $\op$, called the ultimate approximator of $\op$. In more detail,  given a complete lattice $\lat = \langle \set,\leq \rangle$, and $\op : \set \to \set$, the \emph{ultimate approximator} of $\op$, $\uOp : \consSet \to \consSet$ is defined by $\uOp(x,y) = (\bigGlb \op([x,y]) , \bigLub \op([x,y]))$.

\subsection{Semirings}\label{sec:semirings}
In this section we recall the necessary preliminaries on semirings, which we will use as the range of interpretations of logic programs.
A \emph{monoid} is a tuple $\mon = \langle \set, \circ, e \rangle$ such that $\circ: \set \times \set \to \set$ is a binary, associative operator on $\set$;  and we have for any $x \in \set$ that $e \circ x = x = x \circ e$---$e$ is the identity element of $\circ$. If $\circ$ is commutative, we call $\mon$ a \emph{commutative monoid}.
    If $\langle \set,\leq \rangle$ is a poset s.t.\ $\circ$ is $\leq$-monotone, $\mon$ is \emph{ordered} by $\leq$, and if for every $x \in \set$ we have that $e \leq x$,  it is \emph{positively ordered}.

A commutative monoid $\mon = \langle \set, \circ, e \rangle$ is equipped with a 'sum' operation $\bigcirc_I$ for finite index sets $I$ such that:
$\bigcirc_{i \in \emptyset}x_i = e$;
         $\bigcirc_{i \in \{j\}}x_i = x_j$;
         $\bigcirc_{i \in \{j,k\}}x_i = x_j \circ x_k$ for $j \neq k$; and
         $\bigcirc_{j \in J}\bigcirc_{i \in I_j}x_i = \bigcirc_{i \in I}x_i$ if $\bigcup_{j \in J}I_j = I$ and $I_j \cap I_{j^\prime} = \emptyset$ for $j \neq j^\prime$.
    If the sum operation $\bigcirc_I$ is well-defined for any infinite index set $I$, $\mon$ is a \emph{complete monoid}.
    \footnote{Bijective indexings are left implicit. E.g. $\bigcirc \{x_1,\ldots,x_n\} = x_1 \circ \ldots \circ x_n$ and $\bigcirc (x_1,x_2,\ldots) = x_1 \circ x_2 \circ \ldots$ (if it is defined).}

    A \emph{semiring} is a tuple $\sem = \langle \set, +, \times, \0, \1 \rangle$, such that:
$\langle \set, +, \0 \rangle$ is a commutative monoid, with $+$ being called the additive operator;
   $\langle \set, \times, \1 \rangle$ is a monoid, with $\times$ being called the multiplicative operator;
    we have for any $x \in \set$ that  $x \times \0 = \0 = \0 \times x$---i.e. $\0$ is the absorbing element of $\times$;
    and we have for any $x,y,z \in \set$ that $x \times (y + z) = (x \times y) + (x \times z)$ and $(y + z) \times x = (y \times x) + (z \times x)$---i.e. $\times$ left- and right-distributes over $+$.
\emph{Sum} ($\sum$) and \emph{product} ($\prod$) denote the sum operation of $+$ and $\times$ respectively. We identify six specific classes of semirings:
\begin{enumerate}
\item If $\langle \set, \times, \1 \rangle$ is a commutative monoid, $\sem$ is a \emph{commutative semiring}.
\item If $x+x=x$ holds for any $x \in \set$, $\sem$ is an \emph{idempotent semiring}.
\item If $\sem$ is a commutative idempotent semiring, and we have for any $x \in \set$ that $x + \1 = \1 = \1 + x$---i.e. $\1$ is the absorbing element of $+$---then $\sem$ is a \emph{constraint-based semiring (c-semiring)}~\cite{bistarelliSemiringbasedContstraintLogic2001}.
\item $\sem$ is a \emph{complete semiring} if $\langle \set, +, \0 \rangle$ is a complete monoid, and it holds that $\sum_{i \in I}(x \times x_i) = x \times (\sum_{i \in I}x_i)$ and $\sum_{i \in I}(x_i \times x) = (\sum_{i \in I}x_i) \times x$.
\item If $\langle \set,+,\0 \rangle$ and $\langle \set, \times,\1 \rangle$ are ordered by poset $\langle \set, \leq \rangle$, $\sem$ is \emph{ordered} by $\langle \set, \leq\rangle$.
\item If $\langle \set,+,\0 \rangle$ is positively ordered by the poset $\langle \set, \leq\rangle$ and $\langle \set, \times,\1 \rangle$ is ordered by $\langle \set, \leq\rangle$, then $\sem$ is \emph{positively ordered} by $\langle \set, \leq\rangle$.
\end{enumerate}

\begin{note}
Hereafter, unless otherwise specified, $\sem = \langle \set, +,\times,\0,\1 \rangle$ is assumed to be a commutative semiring.
\end{note}
Some examples of commutative semirings are:
\begin{itemize}
    \item $\mathbb{S} = \langle \mathbb{F},+,\cdot,0,1\rangle$, for $\mathbb{F}\in\{\mathbb{N},\mathbb{Z},\mathbb{Q},\mathbb{R}\}$;
\item $\mathcal{P}(A) = \langle 2^A,\bigcup,\bigcap,\emptyset,A\rangle$, the power set semiring ;
    \item $\mathbb{B} = \langle \{\textit{true},\textit{false}\},\lor,\land,\textit{false},\textit{true}\rangle$, the Boolean semiring;
    \item $\mathbb{F} = \langle [0,1], \max, \min, 0,1 \rangle$, the fuzzy semiring;
\item $\mathbb{N}_\infty = \langle \mathbb{N} \bigcup \{\infty\},+,\cdot,0,1 \rangle$, where $n + \infty = \infty$ for all $n$ and $m \cdot \infty = \infty$ for all $m \neq 0$ --- the semiring of natural numbers lifted to infinity.
\end{itemize}
 \section{Families of Semirings and Their Orderings}
\label{sec:families of semirings and their orderings}
In Section \ref{sec:lattice:theory}, we saw a constructive definition for the least fixpoint of a monotone operator on a complete lattice, and, in Section \ref{ssec:general background approximation fixpoint theory}, we saw that such constructions are widely used in AFT.
In the following sections, we will use such constructions again to define the semantics of our semiring-based formalism.
For this, we will require complete lattices that order our semirings.
Some semirings of interest have standard orderings that happen to be complete lattices---for example, the Boolean semiring $\mathbb{B}$ forms a complete lattice under $\textit{false} < \textit{true}$ and the power set semiring $\mathcal{P}(A)$ forms a complete lattice under set inclusion.
Such orderings are not readily apparent for all semirings, however.
We therefore investigate a ``semiring-agnostic'' ordering and look for conditions under which this ordering forms a complete lattice.
This ordering, found throughout the literature~\cite{greenProvenanceSemirings2007,khamisConvergenceDatalogPre2024,hannulaConditionalIndependenceSemiring2023}, is the natural order.

\begin{definition}\label{def:natural order}
    For any $x,y \in \set$, we say $x \leqNat y$ if there exists a $z \in \set$ such that $x+z=y$.
    When $\leqNat$ is a partial order, we call it the \emph{natural order}.
    A semiring $\langle \set, +, \times, \0, \1 \rangle$ for which $\langle \set,\leqNat\rangle$ is a partial order is called \emph{naturally ordered}.
\end{definition}

Not all semirings are naturally ordered.
When semirings allow for additive inverses, the ordering is not antisymmetric.
In all other cases, the ordering is a partial order.
\begin{lemmarep}
    \label{lem:natural order iff no additive inverse}
    $\sem$ is naturally ordered if and only if it contains \emph{no} elements $x$ and $-x$ s.t. $x \neq \0$ and $x + (-x) = \0$.
\end{lemmarep}
\begin{appendixproof}
    We show that ($\dagger$) $\sem$ is naturally ordered if it contains \emph{no} elements $x$ and $-x$ s.t. $x \neq \0$ and $x + (-x) = \0$, and ($\ddagger$) $\sem$ is not naturally ordered if it \emph{does} contain elements $x$ and $-x$ s.t. $x \neq \0$ and $x + (-x) = \0$.

    We begin with the proof for $\dagger$.
    Recall that $\langle \set,\leqNat \rangle$ is a partial order if it is reflexive, transitive, and antisymmetric.
    We show in turn that $\leqNat$ has each of these three properties when $\langle \set, +, \times, \0, \1 \rangle$ is a semiring s.t. $\set$ contains \emph{no} elements $x$ and $-x$ s.t. $x \neq \0$ and $x + (-x) = \0$.
    Note that this assumption---the absence of additive inverses---is only relevant to show antisymmetry.
    \begin{description}
        \item[Reflexivity:] By definition of a semiring, $\0$ is the neutral element for $+$. 
        We thus have for all $a \in \set$ that $a + \0 = a$.
        By the definition of $\leqNat$ this gives $a \leqNat a$ for all $a \in \set$.
        \item[Transitivity:] Let $a,b,c \in \set$ s.t. $a \leqNat b$ and $b \leqNat c$.
        By the definition of $\leqNat$ we know that there exists an $x \in \set$ s.t. $a + x = b$ and there exists a $y \in \set$ s.t. $b + y = c$.
        Substituting $(a + x)$ for $b$ in $b + y = c$ gives $(a + x) + y = c$.
        Associativity of $+$ gives $a + (x + y) = c$.
        We now have that there exists a $z \in \set$ s.t. $a + z = c$ and, as such, that $a \leqNat c$.
        \item[Antisymmetry:] We offer a proof by contraposition, showing that $\set$ contains additive inverses if $\langle \set,\leqNat \rangle$ is not antisymmetric.
        Let $a,b \in \set$ s.t. $a \leqNat b$, $b \leqNat a$, and $a \neq b$.
        By the definition of $\leqNat$ we know that there exists an $x \in \set$ s.t. $a + x = b$ and there exists a $y \in \set$ s.t. $b + y = a$.
        $a + x = b$ and $b + y = a$, together with the assumption that $a \neq b$ gives us that $x \neq \0$ and $y \neq \0$.
        We substitute $a + x$ for $b$ in $b + y = a$ to get $(a + x) + y = a$.
        By associativity of $+$ we get $a + (x + y) = a$.
        From here we see $x + y = \0$.
        $x + y = \0$ and $x \neq \0$, so $\set$ contains additive inverses.
    \end{description}
    This concludes the proof for $\dagger$.

    We proceed with the proof for $\ddagger$.
    Let $\set$ contain $x$ and $-x$ s.t. $x \neq \0$ and $x + (-x) = \0$.
    Because $x + (-x) = \0$, by definition of $\leqNat$, we have that $x \leqNat \0$.
    Because $\0$ is the neutral element of $+$, we have that $\0 + x = x$, and therefore that $\0 \leqNat x$.
    Recall that we assumed that $x \neq \0$.
    We have now found elements $y = x \in \set$ and $z = \0 \in \set$ s.t. $y \leqNat z$, $z \leqNat y$, and $y \neq z$.
    $\leqNat$ is not antisymmetric, and therefore not a partial order.
    This concludes the proof for $\ddagger$.
\end{appendixproof}

Our additive and multiplicative operators are monotone for naturally ordered semirings.
\begin{lemmarep}\label{lem:natural order, monotonicity}
    If $\sem$ is naturally ordered, it is ordered by $\leqNat$.
\end{lemmarep}
\begin{appendixproof}
Let $\langle \set, +, \times, \0, \1 \rangle$ be a semiring with $a, a^\prime, b \in \set$, such that $\langle \set, \leqNat \rangle$ is a partial order.
    We show that $a \leqNat a^\prime$ implies $a + b \leqNat a^\prime + b$ and $a \times b \leqNat a^\prime \times b$.
    From here, commutativity gives us that both operators are monotone.

    We first consider $+$.
    Assume $a \leqNat a^\prime$.
    By the definition of $\leqNat$, this means there exists some $x \in \set$ such that $a + x = a^\prime$.
    We start with $a^\prime + b$.
    Substituting $a + x$ for $a^\prime$, we get $a^\prime + b = (a + x) + b$.
    Using associativity and commutativity, we rewrite this to get $(a + x) + b = (a + b) + x$.
    We now have $(a + b) + x = a^\prime + b$.
    By definition of $\leqNat$, this gives $a + b \leqNat a^\prime + b$.

    We now consider $\times$.
    Once again we assume $a \leqNat a^\prime$ and derive from this that there exists some $x \in \set$ such that $a + x = a^\prime$.
    We start with $a^\prime \times b$.
    Substituting $a + x$ for $a^\prime$ gives $a^\prime \times b = (a + x) \times b$.
    Because $\times$ distributes over $+$, we get $(a + x) \times b = (a \times b) + (x \times b)$.
    We now have $(a \times b) + (x \times b) = a^\prime \times b$.
    Again, by definition of $\leqNat$, this gives $a \times b \leqNat a^\prime \times b$.
\end{appendixproof}

A complete lattice of semiring elements must be bounded.
For naturally ordered semirings, the element $\0$ is the least element bounding our lattice from below.
\begin{lemmarep}\label{lem:natural order, least element}
    If $\sem$ is naturally ordered, $\botNat = \0$.
\end{lemmarep}
\begin{appendixproof}
Let $\langle \set, +, \times, \0, \1 \rangle$ be a semiring s.t. $\langle \set,\leqNat \rangle$ is a partial order.
    Because $\0$ is the neutral element for $+$ we have that $\0 + a = a$ for any $a \in \set$.
    By the definition of $\leqNat$ this gives $\0 \leqNat a$ for all $a \in \set$.
    As $\langle \set,\leqNat \rangle$ is a partial order and therefore antisymmetric, we have for any $b \in \set$ s.t. $b \leqNat \0$ that $b = \0$.
    $\0$ is lesser than all elements, and no other element is lesser than $\0$, so $\botNat = \0$.
\end{appendixproof}

While all naturally ordered semirings are bounded from below by $\0$, they are not generally bounded from above.
A straightforward example of this is the whole number semiring $\mathbb{N}$, for which the natural order coincides with the usual ordering and which extends infinitely upwards.
Naturally, this also means not all naturally ordered semirings form a complete lattice.
In fact, a naturally ordered semiring has a maximal element if and only if said element is absorbing for $+$.
\begin{lemmarep}\label{lem:natural order, maximal element}
    If $\sem$ is naturally ordered, $\topNat = x \in \set$ if and only if $x$ is absorbing for $+$.
\end{lemmarep}
\begin{appendixproof}
    Let $\langle \set, +, \times, \0, \1 \rangle$ be a semiring s.t. $\langle \set,\leqNat \rangle$ is a partial order.
    We show that an element is the maximal element under $\langle \set,\leqNat \rangle$ iff it is absorbing for $+$.

    We first show that maximal element $\topNat \in \set$ is absorbing for $+$.
    $\topNat$ being the maximal element means that for any $x \in \set$ we have $x \leqNat \topNat$ and, as such, that there exists a $y \in \set$ s.t. $x + y = \topNat$.
    Because $\langle \set,\leqNat \rangle$ was assumed to be a partial order, it is also antisymmetric.
    By this antisymmetry, we now have for all $x \in \set$ that if there is a $y \in \set$ s.t. $\topNat + y = x$, i.e. that $\topNat \leqNat x$, then $\topNat = x$.
    Take any $a \in \set$.
    We have that $a + \topNat = \topNat + a$ is some element $b \in \set$.
    Since we found for all $x \in \set$ that if there is a $y \in \set$ s.t. $\topNat + y = x$ then $\topNat = x$, we have that $b = \topNat$.
    Because of this, we conclude that $a + \topNat = \topNat + a = \topNat$ holds for any $a\in \set$, and, thus, that $\topNat$ is absorbing for $+$.

    We now show that any element in $\set$ that is absorbing for $+$ is the maximal element $\topNat \in \set$.
    Take any $a \in \set$ s.t. $a$ is absorbing for $+$.
    This means that for any $b \in \set$ we have that $a + b = b + a = a$.
    Since $b + a = a$ for any $b \in \set$, by definition of $\leqNat$, we have that $b \leqNat a$ for any $b \in \set$.
    This means $a$ is $\topNat$, the maximal element under $\langle \set,\leqNat \rangle$.
\end{appendixproof}

Bistarelli et al.~\citeyear{bistarelliSemiringbasedContstraintLogic2001} propose an ordering specific to c-semirings which forms a complete lattice under certain circumstances.
\begin{definition}\label{def:c-semiring order}
    Let $\sem$ be a c-semiring.
    For any $a,b \in \set$, we define the \emph{c-semiring order} as $a \leqC b$ if $a+b=b$.
\end{definition}

We treat $\leqC$ as if it were defined for any (non-c-)semiring and find it to be a specific case of the natural order.
\begin{lemmarep}\label{lem:orders equivalent}
    For idempotent $\sem$ and $a,b \in \set$: $a \leqNat b$ iff $a \leqC b$.
\end{lemmarep}
\begin{appendixproof}
Let $\langle \set, +, \times, \0, \1 \rangle$ be a semiring such that $+$ is idempotent.
    Take any $a,b \in \set$.

    We first show that $a \leqC b$ implies $a \leqNat b$.
    By $a \leqC b$, we know that $a + b = b$.
    This means that there exists an $x \in \set$, namely $b$, s.t. $a + x = b$.
    By definition of $\leqNat$, we now have $a \leqNat b$.

    We now show that $a \leqNat b$ implies $a \leqC b$.
    By $a \leqNat b$, we know that there exists some $x \in \set$ s.t. $a + x = b$.
    By idempotence of $+$, we get $a + a + x = b$.
    Substituting $b$ for $a + x$ now gives $a + b = b$.
    By definition of $\leqC$, we now have $a \leqC b$.
\end{appendixproof}

Not only do $\leqNat$ and $\leqC$ coincide whenever $\sem$ is idempotent---in this case they are also always a partial order.
This, of course, also means that idempotent semirings do not permit additive inverses.
\begin{lemmarep}\label{lem:natural order with idempotent addition}
    If $\sem$ is idempotent, it is naturally ordered.
\end{lemmarep}
\begin{appendixproof}
    Recall that $\langle \set,\leqNat \rangle$ is a partial order if it is reflexive, antisymmetric, and transitive.
    In the proof for Lemma \ref{lem:natural order iff no additive inverse} we have shown that $\langle \set,\leqNat \rangle$ is reflexive and transitive for any semiring.
    We therefore only show that when $\langle \set, +, \times, \0, \1 \rangle$ is a semiring with idempotent $+$, $\langle \set,\leqNat \rangle$ is antisymmetric.

    Let $a,b \in \set$ s.t. $a \leqNat b$ and $b \leqNat a$.
    By Lemma \ref{lem:orders equivalent}, because $+$ is idempotent, we have that $a \leqC b$ and $b \leqC a$.
    This means, by definition of $\leqC$, that $a + b = b$ and $b + a = a$.
    Through commutativity of $+$ we get $a = b + a = a + b = b$.
    When $a \leqNat b$ and $b \leqNat a$ we get that $a = b$, so $\langle \set,\leqNat \rangle$ is antisymmetric.
\end{appendixproof}

Bistarelli et al.~\shortcite{bistarelliSemiringbasedConstraintSatisfaction1997}  show how $\leqNat$ can form a complete lattice for c-semirings by first showing that $+$ and $\lub$ coincide, then showing that any set that has the $\lub$ also has the $\glb$, and finally combining these two facts to get that any set has both the $\lub$ and $\glb$.
However, this reasoning relies on the assumption that the sum is defined for any (infinite) set of c-semiring elements.
This assumption does not hold for every c-semiring, as shown in Appendix \ref{Appendix:AdditionalResults:Semirings}.
We also find that a complete c-semiring is a stronger assumption than necessary to obtain these results.
Indeed, we find the same results for complete idempotent semirings with a greatest element $\topNat$.

\begin{lemmarep}\label{lem:natural order, complete lattice from complete idempotent and top}
If $\sem$ is a complete idempotent semiring s.t. $\topNat$ exists, then $\langle \set, \leqNat \rangle$ is a complete lattice.
\end{lemmarep}
\begin{appendixproof}
    Working toward a proof for this result, we first demonstrate two intermediate results.
    The first of these intermediate results ($\dagger_1$) is that if $\sem$ is a complete idempotent semiring s.t. $\topNat$ exists, then $\sum X = \bigLub X$ for any $X \subseteq \set$.
    The second of these intermediate results ($\dagger_2$) is that if $\sem$ is a complete idempotent semiring s.t. $\topNat$ exists and $\bigLub X$ exists for any $X \subseteq \set$, then $\bigGlb X$ exists for any $X \subseteq \set$.

    We start with the proof for $\dagger_1$.
    For clarity, we consider empty and non-empty $X \subseteq \set$ as separate cases.
    \begin{description}
        \item[Empty subsets]
            By definition, $\sum \emptyset = \0$.
            By convention, any $x \in \set$ is an upper bound of $\emptyset \subseteq \set$.
            By Lemma \ref{lem:natural order, least element}, we know that $\0 \leq x$ for any $x \in \set$.
            Now, by the definition of \emph{least} upper bound, $\bigLub \emptyset = \0$.
        \item[Non-empty subsets]
        To show that $\bigLub X = \sum X$ for any subset non-empty $X \subseteq \set$, we show that ($\ddagger_1$) $\sum X \leqNat \bigLub X$ and ($\ddagger_2$) $\bigLub X \leqNat \sum X$.
        From there, our conclusion follows through antisymmetry obtained by Lemma \ref{lem:natural order with idempotent addition}.

        We start with the proof for $\ddagger_1$.
        Take any $x \in X$.
        By idempotence, we have that $x + \sum X = \sum X$.
        This gives $x \leqNat \sum X$ by definition of $\leqNat$.
        Since this holds for all $x \in X$, $\sum X$ is an upper bound of $X$.
        We now get that $\bigLub X \leqNat \sum X$, because, by definition, $\bigLub X \leqNat y$ for any upper bound $y \in \set$ of $X$.
        This concludes the proof for $\ddagger_1$.

        We proceed with the proof for $\ddagger_2$, starting from $((\sum X) + (\bigLub X)) \in \set$.
        We rewrite this as $(\sum X) + (\bigLub X) = \sum (\{ x \mid x \in X\}) + (\bigLub X)$.
        By idempotence, this is equal to $\sum \{ x + \bigLub X \mid x \in X\}$.
        By definition, we have $x \leq \bigLub X$ for any $x \in X$.
        By Lemma \ref{lem:orders equivalent} and the assumed idempotence of $+$, we now also have $x \leqC \bigLub X$, and thus that $x + \bigLub X = \bigLub X$ for any $x \in X$.
        Filling this in gives us $\sum \{ x + \bigLub X \mid x \in X\} = \sum \{ \bigLub X \mid x \in X\}$, which by idempotence is equal to $\bigLub X$.
        We have found that $(\sum X) + (\bigLub X) = \bigLub X$, and thus that $\sum X \leqNat \bigLub X$.
        This concludes the proof for $\ddagger_2$.

        $\ddagger_1$ and $\ddagger_2$ combine through antisymmetry to obtain $\bigLub X = \sum X$ for any non-empty $X \subseteq \set$.
    \end{description}
    We have now demonstrated that $\bigLub X = \sum X$ for any $X \subseteq \set$, concluding the proof for $\dagger_1$.

    We proceed with the proof for $\dagger_2$.
    For clarity, we again consider empty and non-empty $X \subseteq \set$ as separate cases.
    \begin{description}
        \item[Empty subsets] By convention, $\bigLub \emptyset = \botNat = \0$ and $\bigGlb \emptyset = \topNat$.
        \item[Non-empty subsets]
        Take any non-empty $X \subseteq \set$ and consider the set of its lower bounds $Y = \{ y \in \set \mid y \leqNat x\ \text{for all}\ x \in X\}$
        Note that this set of lower bounds always contains at least the minimum element $\botNat = \0$, as it precedes any element (Lemma \ref{lem:natural order, least element}).
        We will demonstrate that $\bigLub Y = \bigGlb X$.

        Take any $x \in X$.
        Recall that for any $y \in Y$ we have $y \leqNat x$.
        This means that $x$ is \emph{an} upper bound of $Y$.
        By definition, $\bigLub Y$ (which was assumed to exist) is the \emph{least} upper bound.
        We therefore have that $\bigLub Y \leqNat x$.

        Since $\bigLub Y \leqNat x$ for any $x \in X$, we know that $\bigLub Y \in Y$.
        $\bigLub Y$ is a lower bound of $X$, and all lower bounds of $X$ are lesser than $\bigLub Y$, meaning $\bigLub Y$ is the greatest lower bound $\bigGlb X$.
    \end{description}
    We have demonstrated that, for any $X \subseteq \set$, the existence of $\bigLub X$ implies the existence of $\bigGlb X$, concluding the proof for $\dagger_2$.

    We now proceed with the proof for the main result.
    To demonstrate that $\langle \set, \leqNat \rangle$ is a complete lattice, we must show it to be a partial order, and must show that $\bigLub X$ and $\bigGlb X$ exist for any $X \subseteq \set$.
    By the assumed idempotence of $+$, Lemma \ref{lem:natural order with idempotent addition} gives us that $\langle \set, \leqNat \rangle$ is a partial order.
    The assumed completeness of $\sem$ gives us that $\sum X$ is well-defined for any $X \subseteq \set$.
    From here, $\dagger_1$ gives us that $\bigLub X$ exists and is equal to $\sum X$ for any  $X \subseteq \set$.
    Finally, by the existence of $\bigLub X$, $\dagger_2$ gives us that $\bigGlb X$ exists for any $X \subseteq \set$.
\end{appendixproof}

Completeness of a semiring, by itself, is not sufficient to guarantee a complete lattice of semiring elements.
\begin{lemmarep}
    Not every complete semiring gives rise to a complete lattice under the natural order.
\end{lemmarep}
\begin{appendixproof}
    Consider the semiring $\langle \mathbb{Q}^+\cup\{\infty\}, +, \times, 0,1\rangle$.
    This semiring is complete, which means that $\sum \{x\in \mathbb{Q}^+\cup\{\infty\}\mid x\leq \sqrt{2}\} \in \mathbb{Q}^+\cup\{\infty\}$ (indeed, it is $\infty$).
    However, $\bigGlb  \{x\in \mathbb{Q}^+\cup\{\infty\}\mid x\leq \sqrt{2}\}=\sqrt{2}\not\in  \mathbb{Q}^+\cup\{\infty\}$.
\end{appendixproof}
Other examples of a semiring inducing a complete lattice under the natural order are $\mathbb{N}_\infty$ and $\langle [0,1],+,\cdot,0,1\rangle$. \section{Semiring-based Constraint Logic Programming}
\label{sec:semiring-based constraint logic programming}
We begin our study of semiring-based constraint logic programming, limiting ourselves to the positive fragment of the full formalism that will be presented in Section \ref{sec:generalized negation for semiring-based constraint logic programming}.
In essence, this special case is
 a simplified account of semiring-based constraint logic programs as introduced by Bistarelli et al.~\citeyear{bistarelliSemiringbasedContstraintLogic2001}, in which we exclude variables and functions from our language, leaving such extensions for future work. On the other hand, we lift their assumption of c-semirings and consider general semirings.

\begin{definition}\label{def:psclp}
Given a semiring $\sem = \langle \set, +, \times, \0, \1 \rangle$ and a set of atoms ${\cal A}$, \emph{generalized atoms} are defined as atoms and semiring values. A \emph{clause} is an expression of the form  $H \lpif B_1,\ldots,B_n$ where $H$, the \emph{head} of the clause, is an atom and each $B_i$ appearing in the (possibly empty) \emph{body} of the clause is a generalized atom.
A \emph{positive semiring-based constraint logic program} for $\sem$ (\emph{PSCLP($\sem$)}, for short\footnote{We write PSCLP when the semiring is clear or unimportant.}), is a set of clauses.
$\atoms$ denotes the atoms used in PSCLP($\sem$) $\prog$.
\end{definition}
Notice that Datalog programs~\cite{ceriWhatYouAlways1989} are 
a special case of PSCLPs, namely the programs PSCLP($\mathbb{B}$).

\begin{example}\label{ex:running, introduction}
We consider the following program $\prog$ (adapted from~\cite{bistarelliSemiringbasedContstraintLogic2001}) as a running example.
\begin{lstlisting}[numbers=none]
        $\mathrm{c_{1}}$:    solution(a)     :- path(a,b).
        $\mathrm{c_{2}}$:    solution(a)     :- path(a,c).
        $\mathrm{c_{3}}$:    path(a,b)       :- mass_transit(a).
        $\mathrm{c_{4}}$:    path(a,c)       :- car(a).
        $\mathrm{c_{5}}$:    mass_transit(a) :- train(a).
        $\mathrm{c_{6}}$:    train(a)        :- 2.
        $\mathrm{c_{7}}$:    car(a)          :- 3.
    \end{lstlisting}
    The semiring over which $\prog$ is defined, from which the integer values in the program stem, is the optimization semiring $\mathbb{O} = \langle \mathbb{N} \bigcup \{\infty\}, \min, +, \infty, 0 \rangle$.
    This semiring allows us to optimize over constraints, by assigning each constraint an integer representing the cost of the connection it models, combining constraints by summing their costs, and comparing constraints using the minimum operator.
    We order the semiring with the complete lattice $\langle \mathbb{N} \bigcup \{\infty\}, \geq \rangle$---higher costs are 'lesser than' lower costs.

    The clauses $\mathrm{c_{6}}$ and $\mathrm{c_{7}}$ describe the costs associated with travelling by train or car respectively.
    $\mathrm{c_{5}}$ tells us that train travel is a form of mass transit.
    Since \verb|mass_transit(a)| is only defined by $\mathrm{c_{5}}$, the cost of using mass transit will only depend on the cost of travelling by train.
    The clauses $\mathrm{c_{3}}$ and $\mathrm{c_{4}}$ inform us that mass transit and car travel each allow us to take a different path, and that the costs associated with following these paths will depend on the cost of using their corresponding mode of transport.
    Finally, the clauses $\mathrm{c_{1}}$ and $\mathrm{c_{2}}$ tell us that either path constitutes a solution to our problem, and that we must compare the costs of either path to obtain the optimal solution to our problem.
\end{example}

The semantics of PSCLPs is given in terms of interpretations that assign semiring values to formulas, essentially generalizing classical interpretations by allowing any semiring value as truth value instead of just the Boolean ones.

\begin{definition}\label{def:interpretation}
    An \emph{interpretation} $\interp$ of some PSCLP $\prog$ based on a semiring $\sem = \langle \set, +, \times, \0, \1 \rangle$ is a mapping $\interp: \atoms \to \set$.
We extend interpretations to semiring element $x\in \set$ by $\interp(x)=x$, to conjunctions of formulas (inductively) by $\interp(A,B)=\interp(A) \times \interp(B)$, and to empty conjunctions by $\interp(\emptyset)=\1$.
    
\end{definition}
\begin{note}
    Hereafter, unless otherwise specified, we assume that $\langle \set, \leq \rangle$ is a partial order, and that $\prog$ is a PSCLP($\sem$).
\end{note}

We endow the set of $\prog$'s interpretations $\interps$ with an ordering that respects the partial order of semiring elements.
\begin{definition}\label{def:partial order of interpretations}
    The \emph{partial order of interpretations} $\langle \interps, \preceq \rangle$ is derived from $\langle \set , \leq \rangle$ such that for any $\interp_1, \interp_2 \in \interps$, $\interp_1 \preceq \interp_2$ if $\interp_1 (x) \leq \interp_2 (x)$ for any atom $x \in \atoms$.
\end{definition}

When $\langle \set , \leq \rangle$ is a complete lattice, so is $\langle \interps, \preceq \rangle$.
We now introduce the notion of model for PSCLPs.
\begin{definition}$\interp \in \interps$ is a \emph{semiring model}
of $\prog$ if it holds for every $H \in \atoms$ that $\sum\{\prod_{i=1}^{n} \interp(B_i) \mid H \lpif B_1, \ldots, B_n \in P\} \leq \interp(H)$.
    $\semModels$ denotes the set of semiring models of $\prog$.
\end{definition}
This notion of model deviates slightly from the traditional notion of model.
The traditional notion of model requires only that the head of each clause be interpreted as no lesser than the clause's body---considering every clause independently.
Our notion of model instead considers all clauses defining an atom in unison, requiring that said atom be interpreted as no lesser than the sum of these clauses' bodies.
The difference between these notions stems from the fact that the operators used traditionally for the comparison between clauses (like $\lor$ or $\max$) are idempotent, while the additive operators used for this purpose in our framework are not necessarily.
In this way, our notion of model treats semirings more seriously as a structure for evaluation.\footnote{We defer an investigation of the differences between these notions of model to Appendix \ref{Appendix:AdditionalResults:TraditionalModels}.}

\begin{example}\label{ex:running example, models}
    We illustrate which interpretations are semiring models of our running example program.
    For interpretation $\interp \in \interps$ to be a semiring model of $\prog$, we require that:
    \begin{itemize}
        \item $3 \geq \interp(\mathtt{car(a)})$ and $2 \geq \interp(\mathtt{train(a)})$;
        \item $\interp(\mathtt{train(a)}) \geq \interp(\mathtt{mass\_transit(a)})$;
        \item $\interp(\mathtt{car(a)}) \geq \interp(\mathtt{path(a,c)})$;
        \item $\interp(\mathtt{mass\_transit(a)}) \geq \interp(\mathtt{path(a,b)})$; and
        \item $\min\{\interp(\mathtt{path(a,c)}),\interp(\mathtt{path(a,b)})\} \geq \interp(\mathtt{solution(a)})$.
    \end{itemize}
\end{example}

Given all the interpretations of a PSCLP, we would like to select a single unique model as the representative one.
In logic programming, this representative model is the minimal model~\cite{lloydFoundationsLogicProgramming1987}, obtained as the intersection of each model's sets of true atoms.
Intuitively, this model minimizes truth ascription while still satisfying all clauses.
In other words, it makes true what must be true and nothing else.
Here, we follow a similar approach but require a generalization of the notion of model intersection.
\begin{definition}Given a complete lattice $\langle \set, \leq \rangle$ and $A \in \atoms$, we define the \emph{model intersection} of $M \subseteq \semModels$ as $\inters M (A) = \bigGlb \{\interp(A) | \interp \in M\}$.
\end{definition}

\begin{lemmarep}
\label{lem:semiring model intersection}
    If $\sem$ is ordered by a complete lattice $\langle \set, \leq \rangle$ and $M \subseteq \semModels$, $\inters M$ is a semiring model for PSCLP($\sem$) $\prog$.
\end{lemmarep}
\begin{appendixproof}
    Let $\prog$ be a PSCLP based on a semiring $\sem = \langle \set, +, \times, \0, \1 \rangle$, such that $\langle \set, \leq \rangle$ is a complete lattice on which $+$ and $\times$ are monotone.
    Let $M$ be a set of semiring models for $\prog$.
    We show that the model intersection of $M$ is also a semiring model.
    That is, we show that it holds for every atom $H$ that $\sum\{\prod_{i=1}^{n} \inters M (B_i) | H \lpif B_1, \ldots, B_n \in P\} \leq \inters M (H)$.

    Take any atom $H \in \prog$ and any semiring model $\interp \in M$.
    For every $B_i$ appearing in a clause $H \lpif B_1, \ldots, B_n \in \prog$, by definition of the glb, we have that $\bigGlb \{\interp (B_i) | \interp \in M\} \leq \interp (B_i)$ --- i.e. we have that $\inters M (B_i) \leq \interp (B_i)$.
    By the assumed monotonicity of $+$ and $\times$, this gives us that ($\dagger$): $\sum\{\prod_{i=1}^{n} \inters M (B_i) | H \lpif B_1, \ldots, B_n \in P\} \leq \sum\{\prod_{i=1}^{n} \interp (B_i) | H \lpif B_1, \ldots, B_n \in P\}$ holds for any atom $H \in \prog$ and any semiring model $\interp \in M$.
    
    Since $M$ is a set of semiring models, we have for any $\interp \in M$ and any atom $H \in \prog$ that $\sum\{\prod_{i=1}^{n} \interp (B_i) | H \lpif B_1, \ldots, B_n \in P\} \leq \interp (H)$.
    Through transitivity of $\leq$, we combine this with $\dagger$ to get that $\sum\{\prod_{i=1}^{n} \inters M (B_i) | H \lpif B_1, \ldots, B_n \in P\} \leq \interp (H)$ for any $\interp \in M$.
    That is, we have that ($\ddagger$): $\sum\{\prod_{i=1}^{n} \inters M (B_i) | H \lpif B_1, \ldots, B_n \in P\}$ is a lower bound of $\{\interp (H) | \interp \in M\}$ for any $H \in \prog$.

    By definition, for any atom $H \in \prog$, we have that $\inters M (H) = \bigGlb \{\interp (H) | \interp \in M\}$ is the \emph{greatest} lower bound of $\{\interp (H) | \interp \in M\}$.
    By $\ddagger$ this gives us that $\sum\{\prod_{i=1}^{n} \inters M (B_i) | H \lpif B_1, \ldots, B_n \in P\} \leq \inters M (H)$ for any $H \in \prog$.
    The model intersection of $M$ is also a semiring model.
\end{appendixproof}

We obtain the minimal semiring model by intersecting $\semModels$.
\begin{definition}\label{def:minimal model}
    The \emph{minimal semiring model} of PSCLP $\prog$ is $\minSemModel = \inters \semModels$.
    $\minSemModel$ is the model-theoretic semantics of $\prog$.
\end{definition}
\begin{example}\label{ex:running example, minimal semiring model}
    For our running example, the minimal semiring model gives us the following.
    \begin{itemize}
        \item $\minSemModel(\mathtt{car(a)}) = 3$ and $\minSemModel(\mathtt{train(a)}) = 2$;
        \item $\minSemModel(\mathtt{mass\_transit(a)}) = 2$;
        \item $\minSemModel(\mathtt{path(a,c)}) = 3$;
        \item $\minSemModel(\mathtt{path(a,b)}) = 2$; and
        \item $\minSemModel(\mathtt{solution(a)}) = 2$.
    \end{itemize}
\end{example}

The model-theoretic semantics of a PSCLP lacks a constructive definition.
We therefore work toward an equivalent fixpoint semantics that does have a constructive definition, based on an extension of the well-known immediate consequence operator from Datalog to PSCLPs.
\begin{definition}\label{def:TP}
Given interpretation $\interp \in \interps$ and atom $H\in\atoms$,
$\imCon (\interp) (H) = \sum\{\prod_{i=1}^{n} \interp(B_i) | H \lpif B_1, \ldots, B_n \in \prog\}$.
\end{definition}

The fixpoint semantics of a PSCLP is defined as the least fixpoint of this immediate consequence operator.
This fixpoint can be constructed if $\imCon$ is monotone on a complete lattice.
\begin{lemmarep}\label{lem:immediate consequence, monotonicity}
    If $\sem$ is ordered by $\langle \set, \leq \rangle$, $\imCon$ is $\preceq$-monotone.
\end{lemmarep}
\begin{appendixproof}
    Let $\sem = \langle \set, +, \times, \0, \1 \rangle$ be a semiring, and $\prog$ a PSCLP($\sem$).
    Consider a partial order of semiring elements $\langle \set, \leq \rangle$ and the partial order of interpretations $\langle \interps, \preceq \rangle$ derived from it as per definition \ref{def:partial order of interpretations}.
    Let $\interp_1, \interp_2 \in \interps$ be two interpretations of $\prog$, such that $\interp_1 \preceq \interp_2$.

    Recall the definition of $\imCon$:
    $\imCon (\interp) (H) = \sum\{\prod_{i=1}^{n} \interp(B_i) | H \lpif B_1, \ldots, B_n \in P\}$ for interpretation $\interp$, and atom $H$.
    We work through this definition from the inside out, starting with the evaluations of the elements $B_i$ making up the various clauses defining $H$.
    Because we assumed that $\interp_1 \preceq \interp_2$, we have that $\interp_1(B_i) \leq \interp_2(B_i)$ for any $H \lpif B_1, \ldots, B_n \in P$.
    By the assumed $\leq$-monotonicity of $\times$, this means that for any $H \lpif B_1, \ldots, B_n \in P$ we have that $\prod_{i=1}^{n} \interp_1(B_i) \leq \prod_{i=1}^{n} \interp_2(B_i)$.
    From here, the assumed $\leq$-monotonicity of $+$ gives us that $\sum\{\prod_{i=1}^{n} \interp_1(B_i) | H \lpif B_1, \ldots, B_n \in \prog\} \leq \sum\{\prod_{i=1}^{n} \interp_2(B_i) | H \lpif B_1, \ldots, B_n \in \prog\}$.
    As such, $\imCon (\interp_1) (H) \leq \imCon (\interp_2) (H)$ for any $H \in \prog$.
\end{appendixproof}

Lemma \ref{lem:natural order, complete lattice from complete idempotent and top} now gives us sufficient conditions for the existence of a constructive definition of our fixpoint semantics.
\begin{corollary}
    \label{cor:least fixpoint tp}
    If $\sem$ is ordered by the complete lattice $\langle \set, \leq \rangle$ (e.g.\ when $\sem$ is a complete idempotent semiring s.t. $\topNat$ exists), there is an ordinal $\alpha$ such that $\imCon^\alpha(\botInterp)$ is the least fixpoint of $\imCon$, which coincides with $\imCon$'s least prefixpoint.
\end{corollary}

\begin{example}\label{ex:running example, tp}
    We illustrate $\imCon$ on the program from Example \ref{ex:running, introduction}.
    The immediate consequence operator is instantiated as $\imCon (\interp) (H) = \min\{\sum_{i=1}^{n} \interp(B_i) | H \lpif B_1, \ldots, B_n \in \prog\}$.
The bottom interpretation $\botInterp$ for this semiring maps each semiring element to itself and each atom to the bottom
element $\infty$.

In view of Corollary \ref{cor:least fixpoint tp}, we obtain $\lfp(\imCon)$ stepwise as illustrated in Table \ref{tab:running example, tp}.
A fixpoint is reached at $\interp_4$, as 
$\imCon(\interp_4)=\interp_4$.
\begin{table}[tb]
        \centering
        \begin{tabular}{ l | c c c c }
            \toprule
                                    & $\interp_1$   & $\interp_2$   & $\interp_3$   & $\interp_4$   \\
            \midrule
            \verb|train(a)|         & $2$           & $2$           & $2$           & $2$           \\  
            \verb|car(a)|           & $3$           & $3$           & $3$           & $3$           \\
            \verb|mass_transit(a)|  & $\infty$      & $2$           & $2$           & $2$           \\
            \verb|path(a,c)|        & $\infty$      & $3$           & $3$           & $3$           \\
            \verb|path(a,b)|        & $\infty$      & $\infty$      & $2$           & $2$           \\
            \verb|solution(a)|      & $\infty$      & $\infty$      & $3$           & $2$           \\
            \bottomrule
        \end{tabular}
        \caption{$\imCon$ applied to Ex.\ \ref{ex:running example, tp}, where $\interp_1=\imCon(\botInterp)$ and $\interp_{i+1}=\imCon(\interp_i)$ for $i>0$.}
        \label{tab:running example, tp}
    \end{table}
\end{example}

To show the equivalence between our model-theoretic semantics and our fixpoint semantics, we require that the prefixpoints of $\imCon$ coincide with the semiring models of $\prog$.
\begin{lemmarep}
    \label{lem:semiring models and prefixpoints of tp are generally equivalent}
Let $\sem$ be ordered by the complete lattice $\langle \set, \leq \rangle$.
$\interp \in \interps$ is a semiring model of PSCLP($\sem$) $\prog$ iff $\imCon(\interp) \preceq \interp$.
\end{lemmarep}
\begin{appendixproof}
Let $\sem = \langle \set, +, \times, \0, \1 \rangle$ be a semiring, and let $\prog$ be a PSCLP($\sem$).
    Take any interpretation $\interp$ of program $\prog$.
    We first show that $\interp$ being a semiring model implies $\imCon(\interp) \preceq \interp$.
    Consider any atom $H \in \atoms$.
    Assuming $\interp$ is a semiring model we know that  $\sum\{\prod_{i=1}^{n} \interp(B_i) \mid H \lpif B_1, \ldots, B_n \in P\} \leq \interp(H)$.
    We recall that $\imCon (\interp) (H) = \sum\{\prod_{i=1}^{n} \interp(B_i) | H \lpif B_1, \ldots, B_n \in \prog\}$.
    We now have that $\imCon (\interp) (H) \leq \interp (H)$ for any ground atom $H$.
    By definition, this gives $\imCon(\interp) \preceq \interp$.

    We now show that $\imCon(\interp) \preceq \interp$ implies that $\interp$ is a semiring model.
    Take any $H \in \atoms$.
    Assuming $\imCon(\interp) \preceq \interp$, we know that  $\imCon (\interp) (H) \leq \interp (H)$.
    By definition of $\imCon$, this gives us that $\sum\{\prod_{i=1}^{n} \interp(B_i) | H \lpif B_1, \ldots, B_n \in \prog\} \leq \interp (H)$.
    We now have that $\sum\{\prod_{i=1}^{n} \interp(B_i) | H \lpif B_1, \ldots, B_n \in \prog\} \leq \interp (H)$ holds for any ground atom $H$ and, thus, that $\interp$ is a semiring model.
\end{appendixproof}
From here, we obtain the equivalence between the minimal semiring model of $\prog$ and the least fixpoint of $\imCon$.
\begin{lemmarep}
Let $\sem$ be ordered by the complete lattice $\langle \set, \leq \rangle$.
For PSCLP($\sem$) $\prog$, $\minSemModel = \lfp(\imCon)$.
\end{lemmarep}
\begin{appendixproof}
We recall an important result from Tarski~\cite{tarskiLatticetheoreticalFixpointTheorem1955} which we will use in this proof.
    \begin{theorem}
        \label{thm:tarski lattices}
        Let $\lat = \langle \set,\leq \rangle$ be a complete lattice.
        For any monotone operator $\op: \set \to \set$ we have the following:
        The set of all fixpoints of $\op$ is non-empty and forms a complete lattice under $\leq$.
        In particular, $\lfp(\op)$ exists.
        Additionally, the least prefixpoint of $\op$ is equal to its least fixpoint.
    \end{theorem}
    By definition of minimal semiring model, $\minSemModel = \bigGlb(\{M | M\ \text{is a semiring model for}\ \prog\})$.
    By Lemma \ref{lem:semiring models and prefixpoints of tp are generally equivalent}, we get $\minSemModel = \bigGlb(\{\interp | \imCon (\interp) \preceq \interp\})$.
    By Theorem \ref{thm:tarski lattices}, this is equivalent to the least fixpoint of $\imCon$.
\end{appendixproof}
\section{Generalized Negation for Semiring-based Constraint Logic Programming}
\label{sec:generalized negation for semiring-based constraint logic programming}
As presented up until this point, our formalism can only represent positive expressions.
We can express that one may use mass transit if there \emph{is} an available train connection, but we would have no good way to express that cycling is a good alternative if it is \emph{not} raining.
We now extend our formalism with a notion of negation so that negative expressions may also be represented.
This notion of negation will follow the idea of negation as failure~\cite{clarkNegationFailure1977}, where the negation of an atom may be derived if the atom itself is not derivable.

\begin{definition}\label{def:nsclp}
Given a semiring $\sem = \langle \set, +, \times, \0, \1 \rangle$ and a set of atoms $\mathcal{A}$, \emph{negated atoms} are defined as $\negSem a$ where $a\in \mathcal{A}$.
Atoms or their negation are referred to as \emph{literals}, and \emph{generalized atoms} are literals or semiring elements.
A \emph{normal semiring-based constraint logic program} based on $\sem$ (NSCLP($\sem$) for short\footnote{We write NSCLP when the semiring is clear or unimportant.}) is a set of \emph{normal clauses} of the form  $H \lpif B_1,\ldots,B_n$ where $H$, the \emph{head}, is an atom and each $B_i$ appearing in the (possibly empty) \emph{body} is a generalized atom.\end{definition}

Normal logic programs, of course, interpret $\textit{true}$ and $\textit{false}$ as each other's negation.
We generalize negation in normal logic programs with a notion of negation similar to that used in Gödel logics~\cite{godel1932intuitionistischen}, also used in the semiring-based formalism of Eiter and Kiesel~\cite{eiterWeightedLARSQuantitative2020}.
Here we make use of $\1$ and $\0$, the neutral and absorbing elements of the multiplicative operator $\times$;
if an atom is interpreted with $\0$---and it would thus nullify the interpretation of any conjunction it appears in---we interpret its negation as $\1$, leaving the interpretation of the further conjunction unaffected.
Meanwhile, if an atom is interpreted as anything other than $\0$---and it would thus not nullify the interpretation of the conjunction it appears in---we interpret its negation as $\0$, nullifying the interpretation of any conjunction it appears in.
Take, e.g.\ $\interp_4$ from Example \ref{ex:running example, minimal semiring model}.
We have $\interp_4 (\negSem\;\mathtt{car(a)}) = \infty$, as $\interp_4 (\mathtt{car(a)}) = 3 \neq \infty$.
Meanwhile, for \verb|rain(a)|---an atom without a clause defining it---we have $\interp_4 (\negSem\;\mathtt{rain(a)}) = 0$, as $\interp_4 (\mathtt{rain(a)}) = \infty$. This notion of negation generalizes negation as known from nlps while making only minimal assumptions on the semiring.

\begin{note}
    A semiring permits only a single absorbing element for $\times$---namely $\0$;
    for any $\times$-absorbing element $a$, we have that $a \times \0 = a = \0 = \0 \times a$.
\end{note}

\begin{definition}\label{def:interpretation NSCLP}
    An \emph{interpretation} $\interp$ (Def. \ref{def:interpretation}) is extended to negated literals by
$\interp(\negSem a) = \begin{cases}
        \1 & \text{if } \interp(a) = \0 \\
        \0 & \text{otherwise}
      \end{cases}
    $.
\end{definition}

As is well-known from normal logic programs, introducing negation leads to non-monotonicity of the $\imCon$-operator:
\begin{example}Consider the NSCLP($\mathbb{B}$) $\prog=\{a \lpif \negSem b\}$, where

$\textit{true} > \textit{false}$ as usual.
Now consider two interpretations $\interp_1 = \{a:\textit{true}, b:\textit{true}\}$ and $\interp_2 = \{b:\textit{false}, b:\textit{false}\}$.
    We see that $\interp_2 \preceq \interp_1$.
    Applying $\imCon$ we get $\imCon (\interp_1)(a)=\interp_1(\negSem b)=\textit{false}$, $\imCon (\interp_2)(a)=\interp_2(\negSem b)=\textit{true}$, and $\imCon (\interp_1)(b)=\textit{false}=\imCon (\interp_2)(b)$.
    Note that, because there is no predicate defining $b$, the value assigned by $\imCon$ to $b$ is \textit{false}---the neutral element for $\lor$ resulting from summation over an empty set---under either interpretation.
    Comparing the values assigned to $a$ and $b$ now shows that $\interp_2 \preceq \interp_1$ but $\imCon(\interp_2) \npreceq \imCon(\interp_1)$.
\end{example}

Naturally, this behaviour means that previous results relying on the assumed $\leq$-monotonicity of $+$ and $\times$ cannot be presumed to hold for NSCLPs.
Most importantly, the non-monotonicity of $\imCon$ means that we can no longer use Theorem \ref{thm:cousot and cousot construction} to construct its least fixpoint.
In the following sections, we will circumvent this behaviour by means of AFT.
 \section{Normal Semiring-based Constraint Logic Programming in Approximation Fixpoint Theory}
\label{sec:normal semiring-based constraint logic programming in approximation fixpoint theory}
In the previous section, we saw that we can no longer derive the monotonicity of our immediate consequence operator from the monotonicity of $+$ and $\times$.
This left us without a constructive definition of the least fixpoint semantics for our program.
In this section, we capture our framework in AFT by approximating the immediate consequence operator, thus bestowing NSCLPs with the semantic notions recalled in Section \ref{ssec:general background approximation fixpoint theory}.

\begin{note}
    Hereafter, unless otherwise specified, we assume that $\langle \set,\leq \rangle$ is a complete lattice.
    Note that, by extension, $\langle \interps,\preceq \rangle$ is assumed to be a complete lattice as well.
\end{note}

\begin{table*}[h!]
        \centering
        
 \resizebox{\textwidth}{!}{\begin{tabular}{ l | c c c c c ||   l| c c c c c}
            \toprule
                                    & $\botPrec$    & $A_1$             & $A_2$             & $A_3$             & $A_4$   & & $\botPrec$    & $A_1$             & $A_2$             & $A_3$             & $A_4$             \\
            \midrule
            $\mathtt{rain(a)}$          & $(\infty,0)$  & $(\infty,\infty)$ & $(\infty,\infty)$ & $(\infty,\infty)$ & $(\infty,\infty)$ &
            $\mathtt{train(a)}$         & $(\infty,0)$  & $(2,2)$           & $(2,2)$           & $(2,2)$           & $(2,2)$           \\
            $\mathtt{car(a)}$           & $(\infty,0)$  & $(3,3)$           & $(3,3)$           & $(3,3)$           & $(3,3)$           &
            $\mathtt{bicycle(a)}$       & $(\infty,0)$  & $(\infty,1)$      & $(1,1)$           & $(1,1)$           & $(1,1)$           \\
            $\mathtt{mass\_transit(a)}$  & $(\infty,0)$  & $(\infty,0)$      & $(2,2)$           & $(2,2)$           & $(2,2)$           &
            $\mathtt{path(a,d)}$        & $(\infty,0)$  & $(\infty,0)$      & $(\infty,1)$      & $(1,1)$           & $(1,1)$           \\
            $\mathtt{path(a,c)}$        & $(\infty,0)$  & $(\infty,0)$      & $(3,3)$           & $(3,3)$           & $(3,3)$           & 
            $\mathtt{path(a,b)}$        & $(\infty,0)$  & $(\infty,0)$      & $(\infty,0)$      & $(2,2)$           & $(2,2)$           \\
            $\mathtt{solution(a)}$      & $(\infty,0)$  & $(\infty,0)$      & $(\infty,0)$      & $(3,0)$           & $(1,1)$           \\
            \bottomrule
        \end{tabular}}
        \caption{Application of $\fImCon$ for Example \ref{ex:running example, approximating with four valued immediate consequence}, starting from $\botPrec$.}
        \label{tbl:running example, approximating with four valued immediate consequence}
    \end{table*}
\subsection{Approximating the Immediate Consequence Operator}
We first consider the most precise, \emph{ultimate}, approximator $\uImCon$ of the immediate consequence operator $\imCon$.
\begin{definition}\label{def:ultimate approximator of immediate consequence}
$\uImCon(\interp_1,\interp_2) := (\bigGlb \imCon([\interp_1,\interp_2]) , \bigLub \imCon([\interp_1,\interp_2]))$ for $\interp_1,\interp_2\in\consInterps$.
\end{definition}
While the ultimate approximator is the most precise approximator, its evaluation is rather costly---requiring the evaluation of $\imCon$ for every interpretation in the input approximation, prior to identifying the glb and lub.
This typically results in a higher computational complexity~\cite{deneckerUltimateApproximationIts2004}.
For this reason we also introduce another approximator which is not necessarily as precise, but is more economical to use.

For this second approximator, we generalize the four-valued operator originally introduced by \cite{fittingFixpointSemanticsLogic2002}.
Recall that consistent approximations $(\interp_l,\interp_u)\in \consInterps$ have a lower bound $\interp_l$---which is most conservative in its interpretation---and an upper bound $\interp_u$---which is most generous in its interpretations.
In other words, for any atom $A \in \atoms$, $\interp_l(A)$ is the lowest value of any $\interp(A)$ with $\interp \in [\interp_l,\interp_u]$ and $\interp_u(a)$ the highest.
Now, to approximate $\imCon$, we find a new lower bound taking the same sum-product as we would when evaluating $\imCon$ but---instead of considering a single interpretation---we evaluate generalized atoms without negation using $\interp_l$ (yielding the most conservative values) and evaluate negated atoms using $\interp_u$ (nullifying whenever possible).
The new upper bound is found the same way, switching $\interp_l$ and $\interp_u$. This way of approximating $\imCon$ relies on the additional assumption that $\0$ lies at the bottom of our semiring ordering, such that nullifying an expression cannot increase its value;
this way, using the given upper bound to evaluate negated literals results in lesser values and vice versa.

\begin{definition}\label{def:four-valued approximator of immediate consequence}
    Let $\interp_1,\interp_2 \in \interps$;
    $H \in \atoms$;
    $\fImConL (\interp_1,\interp_2)(H) = \sum \{ \prod_{i=1}^{n} \interp^\prime (\interp_1,\interp_2)(B_i) | H \lpif B_1, \ldots, B_n \in \prog\}$;
    and $\interp^\prime (\interp_1,\interp_2)(B_i) = \begin{cases}
        I_2(B_i) & \text{if } B_i \text{ is of the form } \negSem F\\
        I_1(B_i) & \text{otherwise}
    \end{cases}$.
    We define $\fImCon (\interp_1,\interp_2) = (\interp_3,\interp_4)$ s.t. for all $A \in \atoms$, $\interp_3 (A) = \fImConL (\interp_1,\interp_2)(A)$ and $\interp_4 (A) = \fImConL (\interp_2,\interp_1)(A)$.
\end{definition}

It follows immediately from the results from \cite[Theorem 5.6]{deneckerUltimateApproximationIts2004} that $\uImCon$ is an approximator of $\imCon$.

Somewhat surprisingly,  $\fImCon$ is not generally $\leqPrec$-monotone:

\begin{example}\label{ex:approximating with subzero elements}
    We present an example wherein $\fImCon$ is not $\leqPrec$-monotone.
    
    Take the semiring $\mathbb{Z} = \langle \mathbb{Z}\cup\{-\infty,+\infty\},+,\cdot,0,1\rangle$ consisting of the integers with positive and negative infinity, ordered by $\leq$ as usual.
    Consider an NSCLP($\mathbb{Z}$) consisting of a single clause $p \lpif \negSem q, r$.
    Now consider two approximations $A_1 = (\interp_1,\interp_2)$ and $A_2 = (\interp_1,\interp_3)$ where $\interp_1 = \{p:-1, q:-1, r:-1\}$, $\interp_2 = \{p:1, q:1, r:1\}$, and $\interp_3 = \{p:0, q:0, r:0\}$.
    We see that $A_1 \leqPrec A_2$.
    We now apply $\fImCon$ to $A_1$ and $A_2$ and get $\fImCon(A_1) = (\{p:0, q:0, r:0\}, \{p:0, q:0, r:0\})$ and $\fImCon(A_2) = (\{p:-1, q:0, r:0\}, \{p:0, q:0, r:0\})$.
    Even though $A_2$ is more precise than $A_1$, we do not have that $\fImCon(A_1) \leqPrec \fImCon(A_2)$;
    in fact, $\fImCon(A_1)$ is strictly more precise than $\fImCon(A_2)$.
    $\fImCon$ is not $\leqPrec$-monotone for this NSCLP.
    The reason for this failure of $\leqPrec$-monotonicity is that  the upper bound of $A_2$ being lower (i.e.\ more precise) than that of $A_1$ (namely $0$ instead of $1$ for every atom) causes $\negSem q$ in the rule body to evaluate to $1$, which in turn ``allows'' the truth value of $r$ to determine the truth value of $p$.
    However, as this ``positive'' influence of $r$ means that $p$ attains the negative value $\interp_1(r)=-1$, we obtain a new lower bound for $p$ that is negative and thus lower than the new lower bound obtained in view of $A_1$.
    Thus,  $\fImCon$ will not behave as an approximator when applied to semirings that have elements below the $\0$-element.
 
    Working out the same example for $\uImCon$, we get $\uImCon(A_1) = (\{p:-1, q:0, r:0\}, \{p:1, q:0, r:0\})$ and $\uImCon(A_2) = (\{p:-1, q:0, r:0\}, \{p:0, q:0, r:0\})$.
    Here we do have that $\uImCon(A_1) \leqPrec \uImCon(A_2)$. This is no coincidence, as $\uImCon$ is guaranteed to be an approximator \cite{deneckerUltimateApproximationIts2004}.
\end{example}

Whenever the additive and multiplicative operators are monotone and $\0$ is the minimum element of the semiring ordering, $\fImCon$ is an approximator of $\imCon$:
\begin{lemmarep}$\fImCon$ is an approximator of $\imCon$ if and only if
$\sem$ is positively ordered by the complete lattice $\langle \set,\leq \rangle$.
\end{lemmarep}
\begin{appendixproof}
    Let $\sem = \langle \set, +, \times, \0, \1 \rangle$ be a semiring such that $\langle \set, \leq \rangle$ is a complete lattice, and let $\prog$ be an NSCLP($\sem$).
    We will show that $\sem$ being positively ordered by $\leq$ implies that $\fImCon$ is $\leqPrec$-monotone ($\dagger_1$) and vice versa ($\dagger_2$).
    We will then show that ($\ddagger$), for any $\interp \in \interps$, we have that $\fImCon(\interp,\interp)_1 = \fImCon(\interp,\interp)_2 = \imCon(\interp)$.
    From there, we have, by definition, that $\fImCon$ is an approximation of $\imCon$ if and only if $\sem$ is positively ordered by $\leq$.

    We begin with the proofs for ($\dagger_1$) and ($\dagger_2$) but we first produce an equivalent formulation for the $\leqPrec$-monotonicity of $\fImCon$.
    Firstly, recall that for any $p_1 = (\interp_l^1,\interp_u^1), p_2 = (\interp_l^2,\interp_u^2) \in \biInterps$, we have that $p_1 \leqPrec p_2$ means $\interp_l^1 \preceq \interp_l^2$ and $\interp_u^2 \preceq \interp_u^1$, and that $\fImCon$ being $\leqPrec$-monotone means it holds that $p_1 \leqPrec p_2$ implies $\fImCon(p_1) \leqPrec \fImCon(p_2)$.
    Now recall that $\fImCon$ is defined as $\fImCon (\interp_1,\interp_2) = (\interp_3,\interp_4)$ s.t. for all $A \in \atoms$, $\interp_3 (A) = \fImConL (\interp_1,\interp_2)(A)$ and $\interp_4 (A) = \fImConL (\interp_2,\interp_1)(A)$.
    We combine these facts to get the following:
    $\fImCon$ is $\leqPrec$-monotone whenever $\interp_1 \mapsto \fImConL (\interp_1,\interp_2)$ is $\preceq$-monotone and $\interp_2 \mapsto \fImConL (\interp_1,\interp_2)$ is $\preceq$-antimonotone.
    
    We proceed with the proof for $\dagger_1$.
    Assume that $\sem$ is positively ordered by $\leq$.
    Observe that, for any atom $A \in \atoms$, $\fImConL (\interp_1,\interp_2)(A)$ is calculated by applying $+$ and $\times$ to a collection of semiring values produced by $\interp_1$ or $\interp_2$ as the interpretations of generalized atoms appearing in the bodies of clauses defining the atom $A$.
    Because any \emph{positive} generalized atom appearing in these clause bodies is interpreted by $\interp_1$, it follows from the assumed orderedness of $+$ and $\times$ that $\interp_1 \mapsto \fImConL (\interp_1,\interp_2)$ is $\preceq$-monotone.
    
    $\interp_2$ is responsible for interpreting negated atoms $\negSem A$.
    Recall that $\interp_2(\negSem A) = \1$ if $\interp_2(A) = \0$, and that $\interp_2(\negSem A) = \0$ otherwise.
    Take $\interp_2^1$ and $\interp_2^2$ s.t. $\interp_2^1 \preceq \interp_2^2$.
    Because we assumed $+$ and $\times$ to be \emph{positively} ordered, we have three possible cases for an atom $A \in atoms$:
    (1) $\interp_2^1(A) = \interp_2^2(A) = \0$, (2) $\interp_2^1(A) = \0 < \interp_2^2(A)$, and (3) $\0 < \interp_2^1(A) \leq \interp_2^2(A)$.
    For the first and third case, we have that $\interp_2^1(\negSem A) = \interp_2^2(\negSem A)$.
    For the second case, we have that $\interp_2^1(\negSem A) = \1$ and $\interp_2^2(\negSem A) = \0$.
    Because of the assumed positive order, we have that $\0 \leq \1$.
    We now see that $\interp_2 \mapsto \fImConL (\interp_1,\interp_2)$ is $\preceq$-antimonotone.
    This concludes the proof for $\dagger_1$.

    We now proceed with the proof for $\dagger_2$.
    Assuming that $\fImCon$ is $\leqPrec$-monotone means $\interp_1 \mapsto \fImConL (\interp_1,\interp_2)$ is $\preceq$-monotone and $\interp_2 \mapsto \fImConL (\interp_1,\interp_2)$ is $\preceq$-antimonotone.
    It follows straightforwardly from $\interp_1 \mapsto \fImConL (\interp_1,\interp_2)$ being $\preceq$-monotone, that $+$ and $\times$ are ordered by $\leq$.
    From here, the assumption that $\interp_2 \mapsto \fImConL (\interp_1,\interp_2)$ is $\preceq$-antimonotone combines with the fact that $\interp_2$ interprets negated atoms to give us that ($\ddagger$) for any atom $A$, $\interp_2^1(A) \leq \interp_2^2(A)$ implies that $\interp_2^2(\negSem A) \leq \interp_2^1(\negSem A)$.
    We demonstrate that this, in turn, implies that $+$ is \emph{positively} ordered by $\leq$ (i.e., that for all $x \in \set$, we have $\0 \leq x$) by showing that assuming the contrary (that there exists some $x \in \set$ s.t. $\0 \nleq x$) leads to contradiction.
    Indeed, we assume there is some $x \in \set$ s.t. $\0 \nleq x$.
    This means $\0 \neq \botSem$.
    Because $\leq$ is a complete lattice, we know $\botSem$ to exist, and that $\botSem \leq \0$ (indeed $\botSem < \0$).
    Now take two interpretations and an atom $A_1$ s.t. $\interp_2^1(A_1) = \botSem \leq \interp_2^2(A_1) = \0$.
    We have $\interp_2^2(\negSem A_1) = \1$, and, since $\0 \neq \botSem$, we have $\interp_2^1(\negSem A_1) = \0$.
    By $\ddagger$ this gives us that $\1 \leq \0$.
    Take now some atom $A_2 \in \atoms$ s.t. $\interp_2^1(A_2) = \0 \leq \interp_2^2(A_2) \neq \0$.
    We have $\interp_2^2(\negSem A_2) = \0$ and $\interp_2^1(\negSem A_2) = \1$, which, by $\ddagger$, gives us $\0 \leq \1$.
    This would give us that $\0=\1$, which we know only to be possible in the trivial semiring.
    However, since we know that $\botSem \neq \0$, $\sem$ is not the trivial semiring.
    We have thus reached contradiction, concluding the proof for $\dagger_2$.

    For $\ddagger$, we have the following.
    For any $\interp \in \interps$ we have that
    $\fImCon(\interp,\interp)_1 = \fImCon(\interp,\interp)_2 = \fImConL(\interp,\interp)$
    such that, for any $H \in \atoms$,
    $\fImConL(\interp,\interp)(H)
    = \sum\{\prod_{i=1}^{n} \interp^\prime(\interp,\interp)(B_i) | H \lpif B_1, \ldots, B_n \in \prog\}
    = \sum\{\prod_{i=1}^{n} \interp(B_i) | H \lpif B_1, \ldots, B_n \in \prog\}
    = \imCon(\interp)(H)$.
\end{appendixproof}

We can now construct the KK fixpoints of our two approximators, to approximate all the fixpoints of $\imCon$, thus generalizing the semantics from PSCLPs to NSCLPs:
\begin{corollary}There is an ordinal $\alpha$ such that $\uImCon^\alpha(\botPrec) = \lfp_{\leqPrec}(\uImCon)$.
    If $\sem$ is positively ordered by the complete lattice $\langle \set,\leq \rangle$, the same holds for $\fImCon$.
\end{corollary}
\begin{note}
    If $\lfp_{\leqPrec}(\uImCon)$ and $\lfp_{\leqPrec}(\fImCon)$ are exact, $\imCon$ has only a single fixpoint.
    This is the case in particular if $P$ is positive.
\end{note}

\begin{example}\label{ex:running example, approximating with four valued immediate consequence}
    We illustrate the use of an approximator by constructing the KK fixpoint of $\fImCon$ for our running example, extended with three additional clauses to incorporate negation.
    That is, $\prog=\{ c_1,\ldots, c_{10}\}$ with:
\begin{lstlisting}[numbers=none]
        $\mathrm{c_{8}}$:    solution(a)     :- path(a,d).
        $\mathrm{c_{9}}$:    path(a,d)       :- bicycle(a).
        $\mathrm{c_{10}}$:    bicycle(a)      :- 1, $\negSem$ rain(a).
    \end{lstlisting}
$\fImConL$ is instantiated as $\fImConL (\interp_1,\interp_2)(H) = \min \{ \sum_{i=1}^{n} \interp^\prime (\interp_1,\interp_2)(B_i) | H \lpif B_1, \ldots, B_n \in \prog\}$.
    Table \ref{tbl:running example, approximating with four valued immediate consequence} shows the application of the corresponding $\fImCon$, starting from $\botPrec$, to construct the KK fixpoint $A_4$.

\end{example}

\subsection{Stable Semantics for Normal Semiring-based Constraint Logic Programs}
KK fixpoints of $\uImCon$ and $\fImCon$ approximate all fixpoints of $\imCon$, but might sanction self-supporting reasoning:
\begin{example}\label{ex:self-supporting}
    Consider an NSCLP($\mathbb{B}$) $\prog$ consisting of two clauses: $p \lpif \negSem q$ and $q \lpif q$.
Applying $\fImCon$ gives us $\fImCon (\botPrec) = (\{p:\textit{false},q:\textit{false}\},\{p:\textit{true},q:\textit{true}\}) = \botPrec$; thus
    $\botPrec$ is the KK fixpoint of $\fImCon$, which means that $q$ stays undecided because of the self-supporting rule $q\lpif q$, which also keeps $p$ undecided (in view of $p \lpif \negSem q$).

\end{example}
The only reason for making $q$ $\textit{true}$ in the upper bound of the KK fixpoint in the example above  is the fact that $q$ was made $\textit{true}$ in the upper bound of the least precise approximation. This type of ``because-I-said-so'' reasoning is avoided by using the stable operator and its $\leqPrec$-least fixpoint---the WF fixpoint, which we study in this section.

We  first observe these operators are well-defined:
\begin{corollary}For any $(\interp_1,\interp_2) \in \consInterps$ there are ordinals $\alpha$ and $\beta$ such that 
    $(\bigGlb \imCon([\cdot,\interp_2]))^\alpha(\botInterp) = \lfp(\uImCon_l(\cdot,\interp_2))$,
    $(\bigLub \imCon([\interp_1,\cdot]))^\beta(\botInterp) = \lfp(\uImCon_u(\interp_1,\cdot))$, and the \emph{stable operator} for $\uImCon$ is
    $\stable(\uImCon)(\interp_1,\interp_2) = (\lfp(\uImCon_l(\cdot,\interp_2)),\lfp(\uImCon_u(\interp_1,\cdot)))$. The same holds for $\fImCon$ if $\sem$ is positively ordered by $\langle \set,\leq\rangle$.
\end{corollary}
The fixpoints of these stable operators, called the stable fixpoints of $\uImCon$ and $\fImCon$ respectively, are $\leqVal$-minimal fixpoints of their respective bilattice operator (this is an immediate corollary of Theorem 4 of~\cite{deneckerApproximationsStableOperators2001}) and
they generalize the WF and stable model semantics known from normal logic programs (see Theorem \ref{thm:boleaan:programs:generalized}).

We can now also construct the well-founded fixpoints of our two stable operators.
These fixpoints approximate all stable fixpoints of their respective bilattice operators.
\begin{corollary}\label{cor:well-founded fixpoints}
    There is an ordinal $\alpha$ such that $\stable(\uImCon)^\alpha(\botPrec) = \lfp_{\leqPrec}(\stable(\uImCon))$.
If $\sem$ is positively ordered by $\langle \set,\leq \rangle$, the same is true for $\fImCon$.
\end{corollary}

\begin{example}\label{ex:well-founded and stable fixpoints}
    To illustrate the use of the stable operator, we apply $\stable(\fImCon)$ to the program presented in Example \ref{ex:self-supporting}
We first construct the WF fixpoint, by repeatedly applying $\stable(\fImCon)$, starting from $\botPrec$.
    Denoting interpretations as sets of \emph{true} atoms, we get the following.
    First, $\botPrec = (\emptyset , \{p,q\})$.
    Then, $\stable(\fImCon)((\emptyset , \{p,q\})) = (\{p\},\{p\})$ since $\lfp(\fImConL(\cdot,\{p,q\})) = \{p\}$ and $\lfp(\fImConL(\cdot,\emptyset)) = \{p\}$.
    Finally, $\stable(\fImCon)(\{p\},\{p\}) = (\{p\},\{p\})$ so $(\{p\},\{p\})$ is the WF fixpoint of $\fImCon$.
    Since this WF fixpoint is exact, we also know it to be the only stable fixpoint of $\fImCon$.
\end{example}

Finally, since our operators generalize those known from normal logic programming, and AFT has been shown to faithfully capture all the main semantics from nlp \cite{pelovWellfoundedStableSemantics2007}, we faithfully generalize these main semantics:
\begin{theoremrep}\label{thm:boleaan:programs:generalized}
Given an  NSCLP($\mathbb{B}$), the following hold:
(1) $(\interp_1,\interp_2)$ is a stable fixpoint of $\fImCon$ iff it is a partial stable model according to \cite{przymusinskiWellFoundedSemanticsCoincides1990};
(2) $(\interp_1,\interp_2)$ is the WF fixpoint of $\fImCon$ iff it is the WF model according to \cite{przymusinskiWellFoundedSemanticsCoincides1990}; 
(3) $(\interp_1,\interp_2)$ is a stable fixpoint of $\uImCon$ iff it is a partial stable model according to \cite{deneckerUltimateApproximationIts2004};
(4) $(\interp_1,\interp_2)$ is the WF fixpoint of $\uImCon$ iff it is the WF model according to \cite{deneckerUltimateApproximationIts2004}.
\end{theoremrep}
\begin{appendixproof}
For the first three items, it suffices to show that $\fImCon$, for any  NSCLP($\mathbb{B}$), coincides with the four-valued operator by \cite{przymusinskiWellFoundedSemanticsCoincides1990}, which is immediate in view of the definition of $\mathbb{B}$. The last two items are immediate.
\end{appendixproof}
 \section{Conclusion, in view of Related Work}
\label{sec:related:work}
We have unified existing semantics \cite{bistarelliSemiringbasedContstraintLogic2001,khamisConvergenceDatalogPre2023} for PSCLPs in this paper, and generalized the syntax and semantics to allow for default negation on the basis of AFT, resulting in a family of well-behaved semantics that also generalizes the well-known semantics for nlps.
To our best knowledge only a few other approaches combine semirings and logic programming, which we discuss now.

Firstly, Eiter and Kiesel \citeyear{eiterASPAnswerSet2020} provide a framework that captures many approaches to answer set programs taking into account algebraic constraints. In this work, terms (like $\mathtt{a}$ in $\mathtt{train(a)}$)  are allowed to be interpreted using semiring values and operations.
However, while terms are assigned semiring values, formulas are still interpreted in a discrete way, by assigning the neutral element of $\times$ if it is true while assigning the neutral element of the $+$ if it is true.
Thus, they use a different interpretation of formulas, where semirings are seen as oracles to evaluate the constraints.
\cite{eiterASPAnswerSet2020} show that a host of different systems, e.g.\ integrations of satisfaction modulo theories into ASP \cite{cabalarUniformTreatmentAggregates2020,cabalarASPSemanticsConstraints2020}, can be captured within their framework, and the differences  described above also apply to such systems.

Secondly, there is work combining logic programs with semirings in generalizations of model-counting  \cite{derkinderenSemiringsProbabilisticNeurosymbolic2024,kimmigAlgebraicModelCounting2017}.
Such works differ from ours in that their semantics are obtained by assigning every discrete or Boolean interpretation of a logic program a semiring value, generalizing the ProbLog semantics \cite{raedtProbLogProbabilisticProlog2007}, whereas we use semiring values to build our interpretations.

Future work includes computational complexity, implementations, and applying AFT-based notions such as stratification \cite{vennekensSplittingOperatorAlgebraic2004}, conditional independence \cite{heyninck2024algebraic} and non-determinism \cite{heyninck2024non}. 
\bibliographystyle{named}
\bibliography{BibTex}

\end{document}